\pdfoutput=1

\documentclass[11pt]{article}

\usepackage{ACL2023}

\usepackage{times}
\usepackage{latexsym}

\usepackage[T1]{fontenc}

\usepackage[utf8]{inputenc}

\usepackage{microtype}

\usepackage{inconsolata}

\usepackage{comment}
\usepackage{booktabs}
\usepackage{booktabs, multirow} 
\usepackage{soul}
\definecolor{mygreen}{HTML}{17c013}
\usepackage{graphicx}
\usepackage{subfig}
\usepackage{changepage,threeparttable} 
\usepackage{textcomp}
\usepackage{enumerate}
\usepackage{soul}
\usepackage{svg}

\title{Generating Continuations in Multilingual Idiomatic Contexts}

\author{Rhitabrat Pokharel \and Ameeta Agrawal \\
PortNLP Lab, Department of Computer Science, Portland State University\\
  \texttt{\{pokharel,ameeta\}@pdx.edu}
  }

\begin{document}
\maketitle

\begin{abstract}
The ability to process idiomatic or literal multiword expressions is a crucial aspect of understanding and generating any language. The task of generating contextually relevant continuations for narratives containing idiomatic (or literal) expressions can allow us to test the ability of generative language models (LMs) in understanding nuanced language containing non-compositional figurative text. We conduct a series of experiments using datasets in two distinct languages (English and Portuguese) under three different training settings (zero-shot, few-shot, and fine-tuned). Our results suggest that the models are only slightly better at generating continuations for literal contexts than idiomatic contexts, with exceedingly small margins. Furthermore, the models studied in this work perform equally well across both languages, indicating the robustness of generative models in performing this task.
\end{abstract}


\section{Introduction}

\begin{figure}[t]
    \centering
    \includegraphics[width=.48\textwidth]{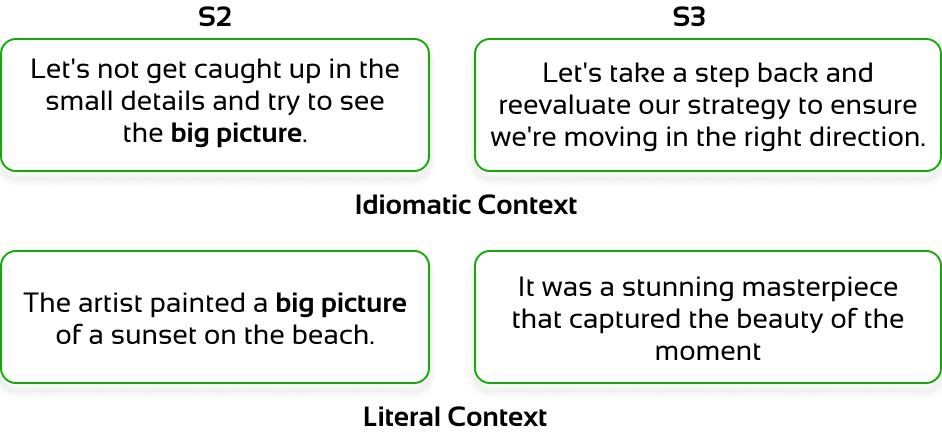}
    \caption{An example where a sentence (S2) contains the same multiword expression used in two contexts -- idiomatic and literal. The task is to generate a coherent follow-up continuation (S3).}
    \label{fig:fig_lit}
\end{figure}

Idiomatic expressions are a common feature of all human languages and are often used to convey emotions, cultural references, and implied meanings. {These are phrases or expressions that have a figurative meaning that is different from the literal meaning of the words that make it up.} In particular, it is the notion of non-compositionality that makes an idiomatic phrase often challenging as it requires understanding the phrase's meaning as a whole. As such, the ability to understand and generate idiomatic expressions is an important task for natural language processing systems, as it allows them to better understand and generate human languages. This is particularly important for applications such as machine translation, language generation, and dialogue systems, where idiomatic expressions are often used to convey meaning. As an example, consider Figure~\ref{fig:fig_lit} where the multiword expression ``big picture'' can convey vastly different meanings depending on the context (idiomatic vs. literal) in which it is being used.

In the field of idiomaticity, prior works have focused on detecting idioms \cite{tayyar2021astitchinlanguagemodels,tan2021does,tedeschi2022id10m,  tedeschi2022ner4id}, paraphrasing idiomatic sentences to literal paraphrases \cite{zhou2022idiomatic},  {cloze task such as {fill-in-the-blank language comprehension}  \cite{zheng2019chid}, classifying idiomatic and literal expressions \cite{peng2015classifying}}, translating idiomatic language  \cite{tang2022petci}, and generating continuations for idiomatic contexts \cite{chakrabarty2022s}.

\begin{table*}[!t]
\centering
\begin{tabular}{p{4.8cm} l p{3cm}}
\toprule
\textbf{Paper} & \textbf{Task} & \textbf{Languages}\\
\midrule
\citet{tayyar2021astitchinlanguagemodels} & Idiomaticity detection & en, pt\\
\citet{tedeschi2022id10m} & Idiomaticity detection & en, de, it, es\\
\citet{tedeschi2022ner4id} & Idiomaticity detection & en, pt, gl\\
\midrule
\citet{tan2021does} & Idioms interpretation & en\\
\citet{chakrabarty2022s} & Idioms interpretation & en\\
\midrule
\citet{moussallem2018lidioms} & Idiom translation, idiom linking & en, de, it, pt, ru\\
\citet{fadaee2018examining} & Idiom translation & en, de\\
\citet{tang2022petci} & Idiom translation & cz, en\\
\midrule
\citet{korkontzelos2013semeval} & Semantic similarity & en, fr, de, it\\
\citet{peng2015classifying} & Idiomatic and literal expression classification & en\\
\citet{zheng2019chid} & Cloze test & cz\\
\citet{chakrabarty-etal-2021-figurative} & Idiomatic continuation generation & en\\
\citet{dashtipour2022extending} & Sentiment analysis of idiomatic sentences & fa\\
\citet{zhou2022idiomatic} & Paraphrasing idioms & en\\
\bottomrule
\end{tabular}
\caption{A survey of works that have focused on idioms in different languages.}
\label{tab:OODist survey}
\end{table*}

The question remains whether generative language models (LMs), typically trained on extensive text corpora of human language, perform differently or similarly under contexts containing literal and idiomatic expressions, particularly in multilingual settings. We explore this by generating text continuations within contexts featuring multiword expressions in both  idiomatic and literal forms. 
Our investigation considers two distinct languages -- English and Portuguese. Both languages use Latin script and subject-verb-object sentence structure. However,  notable  differences exist between these two languages. English is classified as a language with the highest resource level (`5'), whereas Portuguese is categorized as `4' according to the linguistic diversity taxonomy \cite{joshietal2020state}, which could potentially impact how well the models process texts in these languages. Moreover, the distinct traditions and historical influences of Portuguese-speaking and English-speaking cultures lead to differences in social norms and idiomatic expressions.




Using existing datasets of sentence sequences where multiword expressions are used in both literal and idiomatic senses, we empirically evaluate several language models under various settings including zero-shot, few-shot, and fully supervised, by generating logical continuations of narratives. {Our findings suggest that while the models show a slight preference for the literal and compositional use of multiword expressions, resulting in more coherent continuations in literal contexts compared to idiomatic ones, this trend is only consistently observed in approximately half of the cases (with the performance being comparable in the other half). Moreover, the difference is extremely minor, typically not exceeding 0.02 metric points. In terms of multilingual models, our study indicates that all models perform comparably well in both languages,  which is an encouraging outcome. Interestingly, the best results are obtained under the zero-shot setting (rather than few-shot setting) using the GPT-3 \texttt{davinci} model for both English and Portuguese, suggesting that for creative text generation tasks like continuation generation, zero-shot settings are not only effective but also efficient in terms of cost.

The main contributions of this research include: 

\begin{itemize}
    \item Investigating the ability of generative language models to generate coherent subsequent sentences for idiomatic as well as literal contexts; we will make the code\footnote{\url{https://github.com/PortNLP/llm-in-idiomatic-context}} publicly accessible to facilitate further research;
    \item Studying and evaluating four generative models under three training settings (zero-shot, few-shot, and fully supervised) in two distinct languages (English and Portuguese).
    
\end{itemize}




\begin{figure*} [t!]
\centering
\includegraphics[scale=0.6]{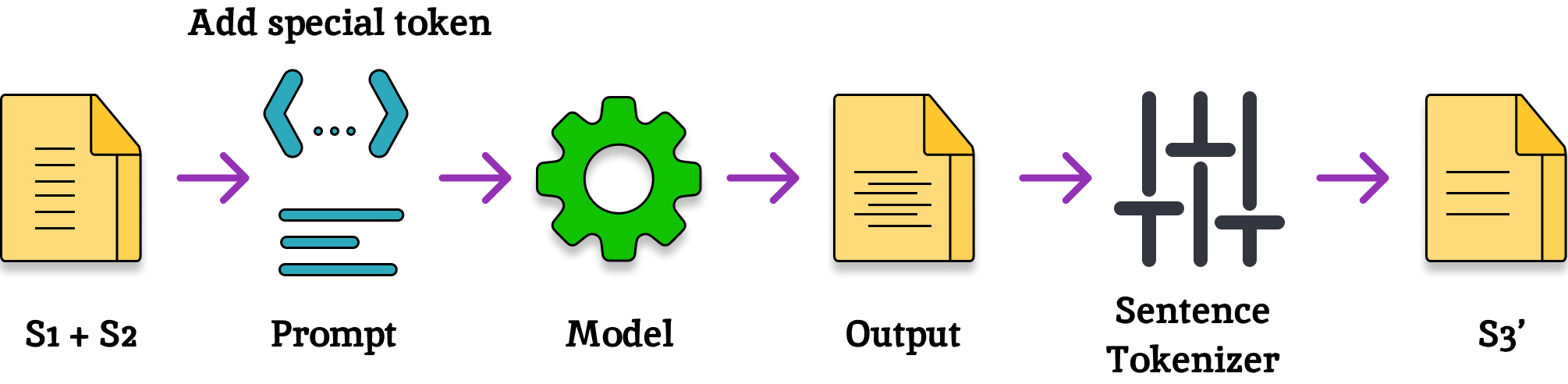}
\caption{Overview of the modeling process.}
\label{fig:framework}
\end{figure*}

\section{Related Work}

{Prior research focusing on idioms can be broadly categorized into two areas: {\em classification} and {\em generative}. Although our work relates to the latter, i.e., generating continuations in multilingual idiomatic contexts, we provide an overview of the background and current developments within both fields of research, and a brief summary in Table~\ref{tab:OODist survey}. In this context, the terms ``idiomatic'' and ``figurative'' are used interchangeably as they both denote language that conveys a meaning that is distinct from its literal or compositional interpretation.} 

\subsection{Idioms-related Classification Tasks}

\citet{tayyar2021astitchinlanguagemodels} studied several transformer-based models such as BERT, XLNet, and XLM-RoBERTa for detection of idiomatic expressions in a sentence as a binary classification task, and additionally, proposed a similarity metric to assess the similarity between idiomatic and non-idiomatic expressions. \citet{tedeschi2022id10m} utilized a BERT-based architecture for idiomatic expression detection, while \citet{tedeschi2022ner4id} measured the similarity between a potentially idiomatic expression and its context to detect idiomatic usage.

In addition to idiom detection, the classification method has also been applied to the comprehension of idioms, encompassing a variety of subjects. One of them is the classification of different sentiments conveyed through idiomatic expressions \cite{dashtipour2022extending}. \citet{jhamtani2021investigating} investigated whether dialogue models are able to handle figurative language usage and concluded that they do not perform well in this area. \citet{tan2021does} evaluated the ability of BERT to understand idioms by selecting the correct paraphrase from a set of options. \citet{liu2022testing} examined models by having them choose the correct metaphorical phrase between two opposite metaphorical phrases, concluding that language models do not make use of context when dealing with metaphorical phrases. In addition, one of the tasks conducted by \citet{chakrabarty2022s} involved the selection of a plausible continuation from two candidate options.

\subsection{Idioms-related Generative Tasks}

In contrast to classification tasks, there has been limited exploration of generative tasks related to idiomatic expressions. \citet{zhou2022idiomatic} used the paraphrasing task to study the ability of models to understand idioms by replacing idiomatic expressions with literal paraphrases. They employed BART model and several metrics to compare the generated text with the reference text.
\citet{chakrabarty2022s} explored the task of generating a coherent next sentence for English idiomatic contexts. 


{
While similar in spirit, there are some notable differences between our work and prior work. \citet{chakrabarty2022s}  exclusively focused on idiomatic usages, whereas our study takes a more comprehensive approach by encompassing and comparing the performance of generative models across {\em both} idiomatic and literal language expressions, which is a novel analysis in this area. {It offers a deeper understanding of how these models interpret idiomatic context. Specifically, it sheds light on whether these models consistently interpret idiomatic phrases in the same manner (either literally or idiomatically), or if their interpretation varies depending on the surrounding context.} Moreover, whereas their work was conducted only in English, our investigation extends its reach to two languages: English (EN) and Portuguese (PT). 
}

\section{Method}

\subsection{Problem Description}
Given a text sequence of two consecutive sentences $S1$ and $S2$, such that $S2$ contains a multiword expression used either in a literal sense or an idiomatic sense, the goal is to generate the next sentence $S3'$ that reasonably and logically continues the narrative and is relevant within the context formed by $S1$ and $S2$. To evaluate the quality of the generated continuation $S3'$, we can either compare $S3'$ to the reference text $S3$ or assess it within the context formed by $S1$ and $S2$.




\subsection{Models}

\medskip
Figure~\ref{fig:framework} presents an overview of the modeling process. {Generative language models are used to generate text by learning patterns and structures from large collections of data, allowing them to generate new, coherent sentences based on the learned patterns.} To generate the $S3'$ sentences, we use three generative language models: GPT-2\footnote{\url{https://huggingface.co/gpt2}} (117M), OPT\footnote{\url{https://huggingface.co/facebook/opt-125m}} (125M), GPT-3\footnote{\url{https://openai.com}} (\texttt{ada} and \texttt{davinci} models), under three training settings: 

\noindent (a) {\em Zero-shot}: using the models without any further training, 

\noindent (b) {\em Few-shot}: {fine-tuning the models using a few examples each from idiomatic and literal contexts (full details in Table~\ref{tab:data_statistics}), and }

\noindent (c) {\em Fully supervised}: fine-tuning the models using the entire training dataset. 

To fine-tune the models (GPT-2 and OPT), we first tokenized the input sentences using the GPT2Tokenizer\footnote{\url{https://huggingface.co/docs/transformers/v4.25.1/en/model_doc/gpt2\#transformers.GPT2Tokenizer}}. We then appended the special token $<|endoftext|>$ at the end of each sample  to ensure that the models could correctly recognize the end of the input text. After the output text was generated, we tokenized it using the NLTK tokenizer \cite{bird2006nltk} and extracted only the first sentence of the generated output as $S3'$ in cases where the models generate more than one sentence. 

For GPT-3 models, we only use few-shot and zero-shot settings with the default settings. As input, we provide the context using $S1$ and $S2$, followed by the prompt:

\smallskip
\noindent ``\texttt{\textbackslash n\textbackslash nQuestion: Generate a logical next sentence.\textbackslash nAnswer:}"

\smallskip
\noindent appended to the end of each context. The generated text was cleaned by removing any HTML tags or trailing white spaces. 


\subsection{Implementation Details}
We experimented with three temperature settings (0.6, 0.8, and 1.0) which control the diversity or randomness of the generated output, with temperature = 1 generating the  most diverse and creative text, and temperature = 0 generating the least diverse text. The GPT-2 and OPT models were trained for 20 epochs, while the GPT-3 models were trained for 4 epochs. We set the learning rate to $2e^{-5}$ and use AdamW optimizer to train the models. The maximum sequence length was set to 400 and the batch size to 16. We used HuggingFace's utility function \texttt{generate}\footnote{\url{https://huggingface.co/docs/transformers/v4.25.1/en/main_classes/text_generation\#transformers.GenerationMixin.generate}} by turning on sampling. {When sampling is turned on, the model generates text by randomly selecting the next word based on its predicted probabilities. This allows for more diverse and creative outputs, as compared to deterministic approaches like greedy decoding.} Since the model does not know when to stop the text generation, we set the generated text's minimum length to 20 and maximum length to 100.




\begin{table}[t!]
   \setlength{\tabcolsep}{12pt}
    \centering
    \begin{tabular}{lcccc}
    \toprule
    ~ & \multicolumn{3}{c}{\textbf{Train}} & \textbf{Test} \\
    \cmidrule{2-4}
        ~  & ZS & FS & Full & ~\\
    \midrule
        \textbf{EN}  &   -     & 87     & 3412 & 364\\ 
    
        \textbf{PT}  &  -     & 53   & 1217 & 238\\ 
    \bottomrule
    \end{tabular}
    \caption{Dataset statistics. The test dataset for a language was the same under all the settings (zero-shot (ZS), few-shot (FS), and fully supervised (Full)).} 
    \label{tab:data_statistics}
\end{table}

\begin{table*}[!h]
\small
\centering
\begin{tabular}{p {0.1\linewidth} p{0.21\linewidth} p{0.21\linewidth} p{0.21\linewidth} p{0.05\linewidth} p{0.05\linewidth}}
\toprule
\textbf{MWE} & \textbf{$S1$} & \textbf{$S2$} & \textbf{$S3$} & \textbf{Label} & \textbf{Lang.}\\
\midrule
{\em night owl} & I explain that a cicada is a locust, while circadian refers to patterns of sleep and wakefulness in relationship to light and darkness. & He has always been a \underline{night owl} and I have always been an early morning person. & If the day comes that I am not up by 5, I am probably seriously ill. Or — as I recently read in someone’s obituary — “not able to do lunch.” & {\em I}  &   EN\\
\cmidrule{2-6}
{\em night owl} & However, you need the internet for the remote access features (no monthly fees for remote viewing). & The \underline{Night Owl} system is a good option for small retail or service businesses. & Reolink Eight Channel PoE Video Surveillance System & {\em L}  &   EN\\

\midrule
{\em coração partido} & Fiz isso, inclusive, na exibição do último episódio da série, quando era editor da Rolling Stone. [{\em  I did this during the airing of the last episode of the series, while I was editor of Rolling Stone.}] & Li o resumão (era contra até então), fiz um textão completamente desacreditado pelo que virou a minha profissão e de \underline{coração partido} pelo episódio mequetrefe. [{\em I read the summary (I was against it until then) and wrote a longish response completely disillusioned with what my profession had become and heartbroken by the mediocre episode.}] & O final era estranhamente confuso, talvez condizente com o que vinha acontecendo na série. [{\em The finale was oddly confusing, though perhaps in line with what had been happening in the series.}] &  {\em I}    &   PT\\
\cmidrule{2-6}
{\em coração partido} & Isso ocorre pois os altos índices de estresse provoca aumento da frequência cardíaca, pressão arterial mais alta, coloca mais pressão no coração e prejudica o sistema imunológico. [{\em This occurs because the high stress levels bring about elevated heart rate and higher blood pressure, increase the load on the heart and damage the immune system.}] & Se você sofre de Síndrome do \underline{Coração Partido}, parte do seu órgão aumentará temporariamente e não conseguirá bombear sangue tão bem quanto antes. [{\em If you suffer from Broken Heart Syndrome, part of your heart will temporarily become enlarged and be unable to pump blood as well as it could before.}] & Enquanto isso, o restante do coração continuará trabalhando normalmente ou será exigido um esforço dobrado. [{\em Meanwhile, the rest of the heart will continue to work normally, or it will require extra effort.}] &  {\em L}    &   PT\\

\bottomrule
\end{tabular}
\caption{A few samples from the English and Portuguese training sets. In this table, we include the translations of Portuguese samples only for the sake of enhanced interpretation but these are not part of the dataset. Labels {\em I} and {\em L} indicate the presence of a multiword expression in $S2$ used in an idiomatic or literal sense, respectively.}
\label{tab:data_samples}
\end{table*}

\section{Evaluation}

\subsection{Datasets}
\label{datasets}

We use an exiting dataset called Multilingual Idiomaticity Detection and Sentence Embedding dataset\footnote{\url{https://github.com/H-TayyarMadabushi/SemEval_2022_Task2-idiomaticity}} \cite{tayyar2021astitchinlanguagemodels}. Specifically, we use the English and Portuguese subsets of the data which were collected by a team of 12 judges from naturally occurring sources. The dataset contains sequences of three consecutive sentences with the middle sentence $S2$ containing multiword expressions in either idiomatic or literal sense. Note that this dataset describes these multiword expressions as {\em potentially idiomatic expressions} (PIE), which means $S2$ contains PIEs, which may or may not necessarily be idioms. However, this is the only available dataset that is closest to the task at hand and includes data from two languages. Table~\ref{tab:data_statistics} presents the dataset's statistics, and some sample instances are shown in Table \ref{tab:data_samples}.  In the test data\footnote{We consider the development set from the original dataset as the test data in our experiments as we did not have access to the ground truth labels for the test set.}, the number of idiomatic and non-idiomatic instances was balanced using random undersampling.






\begin{table*}[!t]\centering
\setlength{\tabcolsep}{10.5pt}
\begin{tabular}{cllcc|cc|cc}\toprule
\multirow{2}{*}{\textbf{Lang.}} &\multirow{2}{*}{\textbf{}} &\multirow{2}{*}{\textbf{Model}} &\multicolumn{2}{c}{\textbf{ROUGE-L}} &\multicolumn{2}{c}{\textbf{METEOR}} &\multicolumn{2}{c}{\textbf{BERTScore}}  \\
\cmidrule{4-9}

& & &\textbf{I} &\textbf{L} &\textbf{I} &\textbf{L} &\textbf{I} &\textbf{L} \\
\midrule

\multirow{10}{*}{EN} &\multirow{3}{*}{ZS} &GPT2 &\textbf{0.10} &0.09  &\textbf{0.11} &0.10 &0.55 &{0.55}  \\

& &OPT &{0.10} &0.10 &0.11 &\textbf{0.12} &{0.55} &0.55  \\

& &GPT3 \texttt{ada} &0.11 &\textbf{0.12}  &0.11 &\textbf{0.13} &0.55 &{0.55}  \\

& &GPT3 \texttt{davinci} & {0.12} &\textbf{\underline{0.13}}*  &0.12 &\textbf{\underline{0.14}}* &{0.59} &\textbf{\underline{0.60}}*  \\

\cmidrule{2-9}

& \multirow{3}{*}{FS} &GPT2 &0.10 &{0.10}   &0.10 &\textbf{0.11} &0.53 &\textbf{0.54} \\

& &OPT &0.09 &\textbf{0.10}  &0.11 &{0.11} &0.55 &\textbf{\underline{0.56}} \\

& &GPT3 \texttt{ada} &0.10 &{0.10}  &0.13 &{0.13} &0.52 &\textbf{0.53}  \\

& &GPT3 \texttt{davinci} &0.10 &\textbf{\underline{0.11}}  &\textbf{\underline{{0.14}}} &0.13 &0.54 &\textbf{0.55}  \\

\cmidrule{2-9}

& \multirow{3}{*}{Full} &GPT2 &{0.10} &0.10  & \underline{0.13} & \underline{0.13} &0.53 &{0.53}  \\

& & OPT &0.10 &\textbf{\underline{0.11}}  &0.12 &{0.12} &{\underline{0.55}} &\underline{0.55}  \\

\midrule
\midrule

\multirow{10}{*}{PT} &\multirow{3}{*}{ZS} &GPT2 &0.07 &{0.07}  &0.08 &{0.08} &0.50 &\textbf{0.52}  \\

& &OPT &0.10 &\textbf{0.11}  &\underline{0.12} &{\underline{0.12}}* &0.56 &\textbf{0.57}  \\

& &GPT3 \texttt{ada} &0.06 &{0.06}  &0.07 &{0.07} &0.51 &\textbf{0.52} \\

& &GPT3 \texttt{davinci} &\textbf{\underline{{0.12}}}* &0.11  &\textbf{0.11} &0.10 &{0.60} &\textbf{\underline{0.61}}*  \\

\cmidrule{2-9}

& \multirow{3}{*}{FS} & GPT2 &{0.08} &0.08   &{0.09} &0.09 &0.52 &{0.52} \\

& & OPT &0.10 &\textbf{0.11}   &\underline{0.11} &\underline{0.11} &0.58 &{0.58} \\

& &GPT3 \texttt{ada} &0.09 &\textbf{0.10} &0.08 &{0.08} &0.56 &\textbf{0.58}  \\

& &GPT3 \texttt{davinci} &0.11 &\textbf{\underline{0.12}}  &{0.10} &0.10 &0.58&{{0.58}}  \\

\cmidrule{2-9}

&  \multirow{3}{*}{Full} &GPT2 &0.09 &\textbf{0.10}   &0.11 &{\underline{0.11}} &0.54 &\textbf{0.55} \\

& &OPT &0.10 &\textbf{{0.11}}  &0.11 &{0.11} &0.57 &\textbf{\underline{0.59}}  \\
\bottomrule
\end{tabular}
\caption{Performance of the models for different metrics with temperature set to 1.0. I = Idiomatic, L = Literal, ZS = Zero Shot, FS = Few Shot, Full = Fully finetuned. The higher score between idiomatic and literal comparison is shown in {\bf bold}, for each metric the best result for each training setting is \underline{underlined}, and for each metric the best overall result for each dataset is shown with an *asterisk (where multiple best overall results exist, the one in the more cost-effective setting is shown). The differences between idiomatic and literal scores are found to be {\em not} statistically significant, with $p$-values > 0.4 using $t$-test.}
\label{tab:results}
\end{table*}

\subsection{Metrics}
We conduct automatic and human evaluations of the generated continuations.  For automatic evaluation, we use the following three metrics which compare the generated sentence $S3'$ with a reference sentence $S3$ that is already available in the dataset. 

\begin{itemize}
    \item \textbf{ROUGE-L} \cite{lin2004rouge}, typically used to compare machine-generated text with human reference text, measures the longest common subsequence between the two texts.
    \item \textbf{METEOR} \cite{banerjee2005meteor} is another widely used evaluation metric that aims to measure the degree of lexical and phrasal overlap between a machine-generated text and one or more reference texts.
    \item \textbf{BERTScore} \cite{zhang2019bertscore} {is a semantic similarity metric that uses cosine similarity between the sentence embeddings to compare the meaning of two sentences}. The embedding model we used was \texttt{microsoft/deberta-xlarge-mnli} \cite{he2021deberta}.
\end{itemize}

While the automatic evaluation measuring the similarity between $S3'$ and an existing $S3$ serves as a quick and cost-effective method of evaluation, it may not comprehensively capture the nuances of natural language, particularly when several valid outputs are possible. Therefore, we complement our evaluation by obtaining human assessment of the outputs where $S3'$ is evaluated within the contexts formed by $S1$ and $S2$.





\section{Results and Discussion}

The results of our experiments are evaluated automatically, through human assessment, and qualitatively, as discussed next.

\subsection{Automatic Evaluation}
Table~\ref{tab:results} presents the main results of our experiments, from which we make some observations to answer the following questions.

\medskip
\noindent \textbf{Are literal contexts easier for language models than idiomatic contexts?}
Overall, in both the language datasets and all three metrics, the literal continuations obtain slightly higher scores than idiomatic continuations. However, in looking closely, we observe that the lexical continuations are better than idiomatic continuations in only about half the scenarios or less (11/20, 4/20, and 12/20 for ROUGE-L, METEOR, and BERTScore, respectively). When we consider the absolute difference in performance, it is interesting to note that the lexical continuations are superior to idiomatic continuations only by a very small margin (maximum difference of 0.01, 0.02, and 0.02 points for ROUGE-L, METEOR, and BERTScore, respectively). The results of statistical significance testing ($t$-test) yield $p$-values > 0.4, indicating that the disparities between idiomatic and literal results lack statistical significance. Taken together, these results lead us to conclude that the generative language models process these distinct contexts somewhat similarly, and that idiomatic contexts are not necessarily more challenging than literal contexts in this task.

\begin{figure}[t]
    \centering
    \subfloat{\includegraphics[width=.45\textwidth]{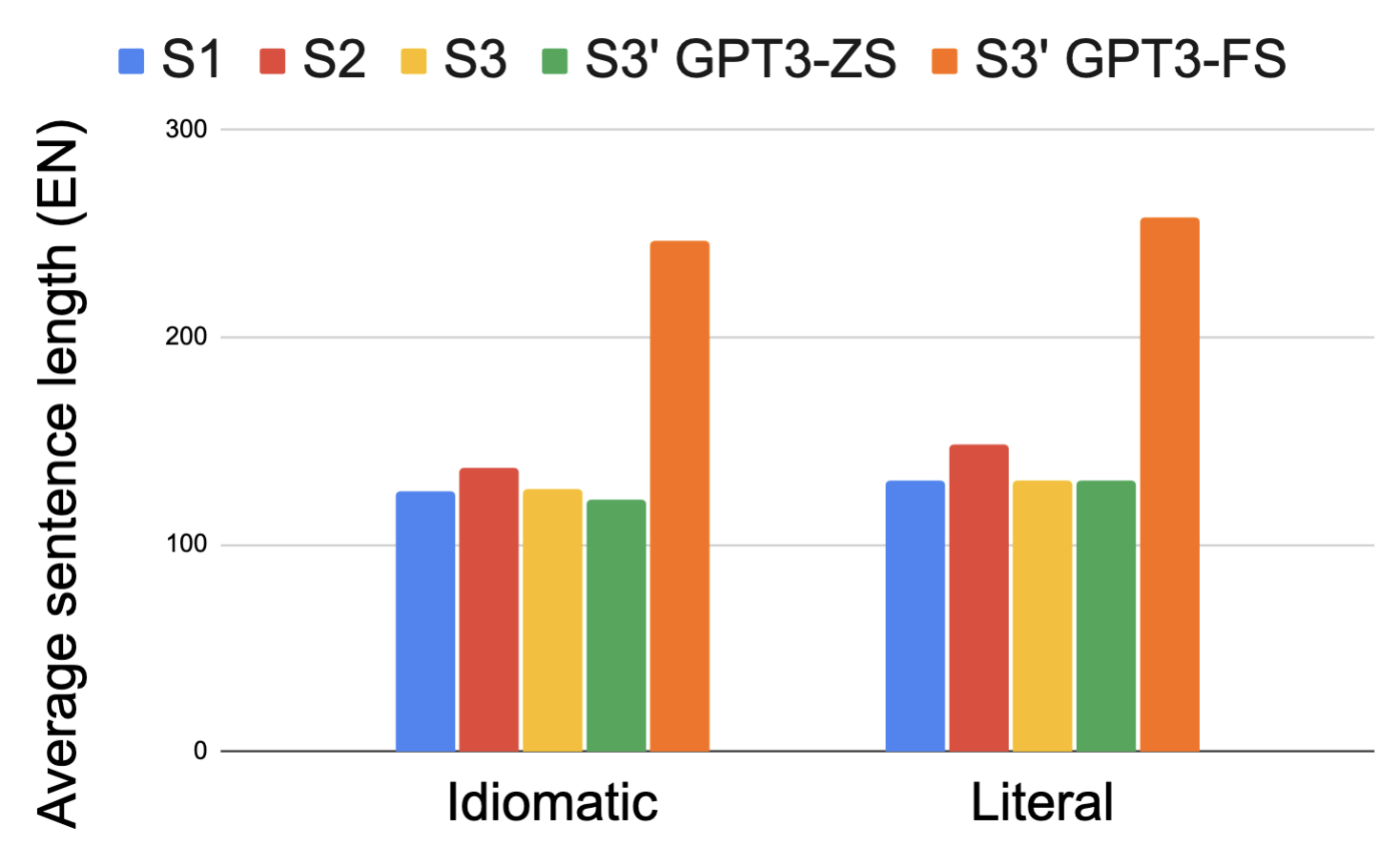}}
    \\
    \subfloat{\includegraphics[width=.45\textwidth]{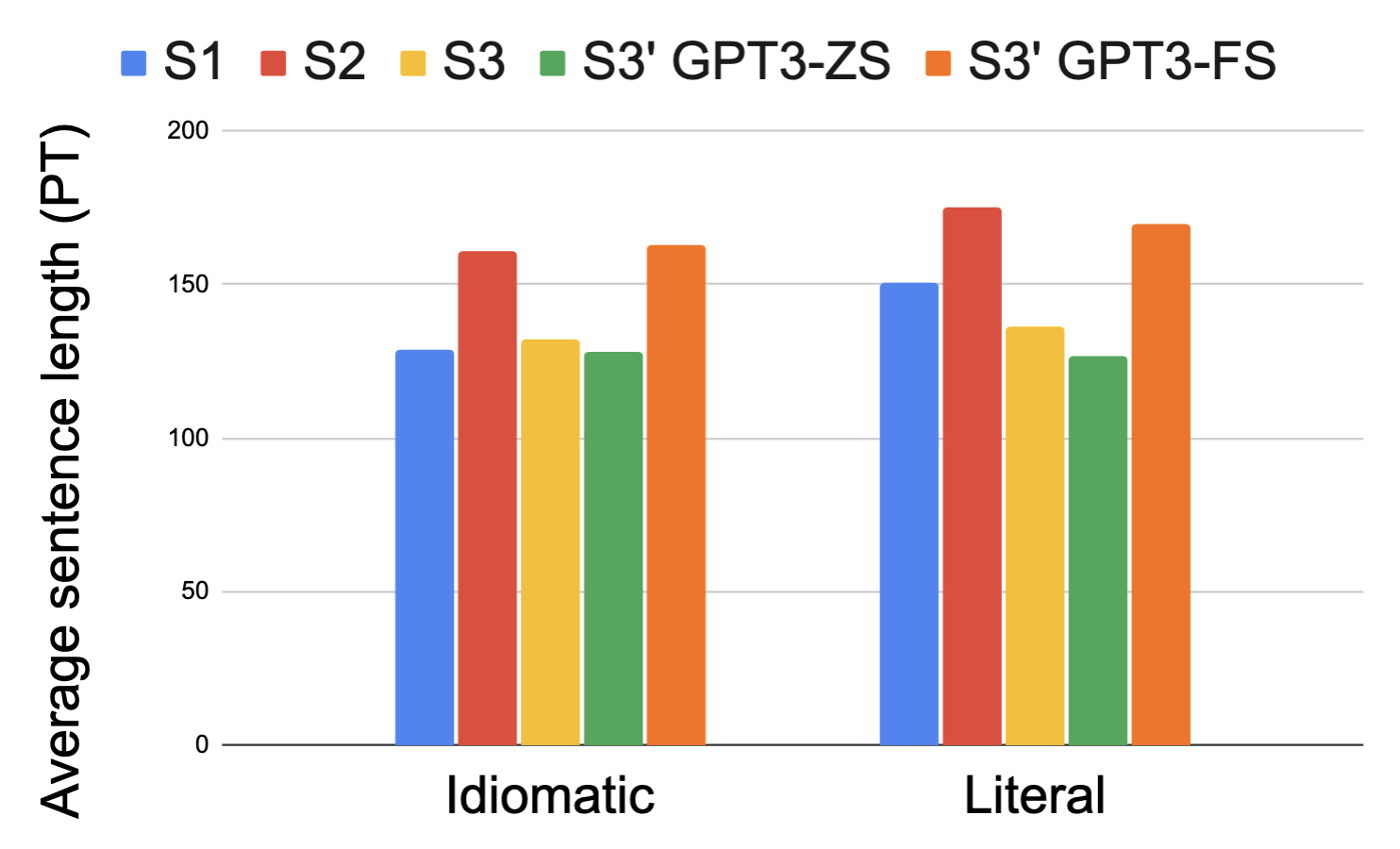}}
    
    \caption{The graph comparing the average lengths of the sentences (numbers of words) for English (top) and Portuguese (bottom).}
    \label{fig:len_comp}
\end{figure}

\begin{figure}[h!]
    \centering
    \subfloat{\includegraphics[width=.33\textwidth]{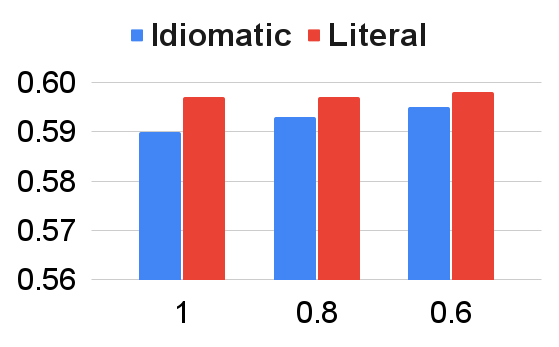}}
    \\
    \subfloat{\includegraphics[width=.33\textwidth]{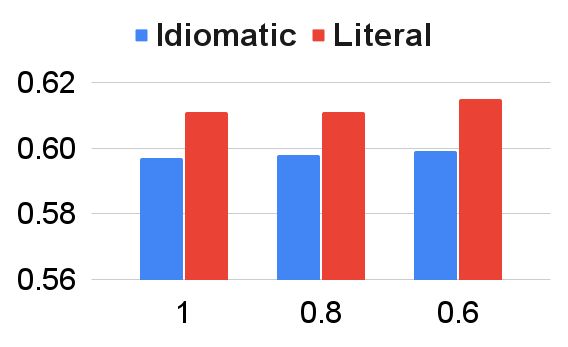}}
    \caption{The results (BERTScore) of GPT-3 \texttt{davinci} under zero-shot for different temperature settings for English (top) and Portuguese (bottom).}
    \label{fig:temp_en_pt}
    \vspace{-0.2cm}
\end{figure}

We analyze the lengths of the different context sentences (Figure~\ref{fig:len_comp}). It is observed that the lengths of $S1$, $S2$, and $S3$ are comparable between the idiomatic and literal contexts. Moreover, in both contexts, $S3'$ generated under the zero-shot setting is similar in length as the original $S3$, while $S3'$ under the few-shot setting is slightly longer. Furthermore, consistent results are obtained under all three temperature settings studied (Figure~\ref{fig:temp_en_pt}).


\medskip
\noindent \textbf{How do language models compare between English and Portuguese?}
In terms of comparing the performance of all LMs between the two different languages, it appears that the results are comparable, which is encouraging given that English is considered the highest resource language (level `5') whereas Portuguese is  `4', a high resource level, in the taxonomy of linguistic diversity \cite{joshi-etal-2020-state}. For all the metrics, performance on English dataset is superior to that of Portuguese dataset by a maximum of 0.05 metric points, and in cases where Portuguese set performs better than English set, it is with at most about 0.04 points, suggesting that the performance across both languages remains largely similar.


\medskip
\noindent \textbf{How do the models perform across different training settings?}
In line with general expectations, the newer and larger model (GPT-3 \texttt{davinci}) generally outperforms the older and smaller models (GPT-2, OPT, GPT-3 \texttt{ada}), even with  no training (zero-shot) or little training (few-shot), although the difference remains small. In comparing the freely available models such as GPT-2 and OPT, a few interesting results emerge: (i) OPT generally outperforms GPT-2 across all settings, but more clearly in Portuguese, (ii) these models benefit from some training especially in the case of Portuguese, and (iii) for English, zero-shot setting yields better results than few-shot setting, but for Portuguese, few-shot setting yields better results than zero-shot setting.

\begin{table}[!t]
\setlength{\tabcolsep}{8pt}
\centering
\small
\begin{tabular}{l|rr|rr}
\toprule
\multirow{2}{*}{\textbf{}}   &\multicolumn{2}{c}{\textbf{METEOR}} &\multicolumn{2}{c}{\textbf{BERTScore}} \\\cmidrule{2-5}
 &\textbf{I} &\textbf{L} &\textbf{I} &\textbf{L} \\
 \midrule
\textbf{Only $S2$ is used}  & & & &\\
\midrule
EN  &0.10 &\textbf{0.11} &0.58 &\textbf{0.59} \\
PT  &\textbf{0.09} &0.08 &0.59 &\textbf{0.61} \\
\midrule
\textbf{$S1$ and $S2$ are used}  & & & &\\
\midrule
EN  &0.12 &\textbf{0.14} &0.59 &\textbf{0.60} \\
PT  &{0.10} &0.10 &0.59 &\textbf{0.61} \\
\bottomrule
\end{tabular}
\caption{Performance of GPT-3 \texttt{davinci} model under zero-shot setting when only $S2$ is used (without $S1$). `I' denotes idiomatic contexts where `L' denotes literal contexts. As comparison, we also add the corresponding results here, borrowing from Table 4.} 
\label{tab:gpt3_davinci_zs}
\end{table}


\begin{table*}[!htp]
\centering
\small
\begin{tabular}{p {0.2\linewidth} p{0.2\linewidth} p{0.2\linewidth} p{0.2\linewidth} p{0.05\linewidth}}
\toprule
\textbf{S1} &\textbf{S2} &\textbf{S3} &\textbf{S3'} &\textbf{Label} \\
\midrule
He believes Mueller’s report on Russian interference in the 2016 election is devastating, but too complicated for mass consumption. &Attorney General William P. Barr spins up a \underline{dust storm} of inaccuracies while Mueller “chose to submit his unreadable — and unread — report and then go away without a sound,” Comey writes. &He could have found a way to speak to the American people in their language. &Mueller's report favours the idea that US intelligence agencies were not effectively informed of the backdrop of Russian activity before the election, and Syria is details which could support or prove the theory's inches. & \textit{I} \\
\midrule
Plant-based food has become a part of the mainstream culinary world.& It’s exciting to be leading this conversation at \underline{Flower Child}.& We work hard to maintain consistency and quality throughout the seasons. &  We believe that everyone, regardless of their diet preference or lifestyle, should have options for amazing plant-based food that not only tastes great but is also good for you.”$\sim$$\sim$\#$\sim$$\sim$POSITION: We are currently looking for an experienced dynamic Vice President of Engineering to join our team.https& \textit{L} \\
\bottomrule
\end{tabular}
\caption{{Qualitative analysis of output generated ($S3'$) by GPT-3 \texttt{davinci}  under zero-shot setting.}} 
\label{tab:generated_text}
\end{table*}

\medskip
\noindent \textbf{How is the performance under limited context?} 
As further analysis, we modify our experimental set up to use only $S2$ as the input context (instead of both $S1$ and $S2$). The results in Table~\ref{tab:gpt3_davinci_zs} show that, as expected, the results are generally lower when only $S2$ is provided. However, this gap is noticeably larger in English than in Portuguese, suggesting that additional contexts are more useful in English than in Portuguese.

\medskip

\subsection{Human Evaluation} 
{For conducting the human evaluation of the generated outputs, two annotators were invited to estimate the relevance and grammatical accuracy of the third sentence ($S3'$) in the context of first ($S1$) and second ($S2$) sentences across 25 randomly selected English samples (12 idiomatic and 13 literal samples) generated from GPT-3 \texttt{davinci} model. 

The annotators were assigned two tasks. 

\medskip
{\bf Task 1} involved rating the relevance of $S3'$ on a scale of 0 to 2, with 0 indicating no relevance, 1 representing neutrality, and 2 signifying relevance. The annotators reached an agreement on 15 samples, which accounts for approximately 60\% of the total. For these 15 samples, both annotators assigned the same relevance scale. Within this subset, 9 samples (about 60\%) were idiomatic, indicating a consistent interpretation across both idiomatic as well as literal contexts by both annotators. Additionally, within this subset, the majority of samples labeled as relevant were idiomatic (7 out of 8). This observation suggests that the model's generated idiomatic continuations were generally preferred.

Overall, considering all the 50 annotations (25 each per annotator), the annotators marked a total of 26 samples (52\%) as relevant (16 idiomatic and 10 literal), 21 (42\%) as neutral (5 idiomatic and 16 literal), and 3 (0.06\%) as not relevant at all (3 idiomatic). These findings indicate that GPT-3 performed well in generating relevant continuations across both the contexts, but particularly so for idiomatic cases.

\medskip
{\bf Task 2} involved identifying any grammatical errors in the generated outputs. These errors primarily included instances where $S3'$ failed to form complete sentences or had some punctuation issues. Other errors included missing spaces after sentence endings, unexpected numbers or symbols inserted into the text, random dates appearing, sentences with unclear or nonsensical content, or unexpected underlined sections. 
{45 out of 50 annotations were flagged as having some kind of abovementioned grammatical errors to some degree and the errors were distributed almost equally between the idiomatic and literal samples. In addition to highlighting the importance of human assessment in natural language generation tasks such as this one, these results suggest that natural language generation continues to present a  challenge for these models.} 
}

\subsection{Qualitative Analysis}
The evaluation of generative tasks, such as narrative continuation, often benefits from qualitative investigation. In this regard, Table~\ref{tab:generated_text} presents a selection of texts generated by the GPT-3 \texttt{davinci} model. It demonstrates that $S3'$ is a logical sentence when considered within its context. However, one can observe certain grammatical errors in the generated text, which contribute to the inconsistency in the results obtained from automated metrics.


\section{Conclusion}

In this work, we investigate the ability of generative language models to generate reasonable continuations under idiomatic and literal contexts. The results suggest that literal continuations seem less challenging for the models than idiomatic continuations, but only slightly so. In particular, the human annotators found the continuations in idiomatic contexts to be fairly relevant. These observations were consistent across English and Portuguese datasets. The GPT-3 \texttt{davinci} model consistently outperformed all other models, and, interestingly, its performance under a zero-shot setting was better than under a few-shot setting.





We have multiple directions for future work that we intend to explore. For example, in this work, we experimented with only a handful of prompts. There are several ways in any language to write the same prompt. As such, the generated text might depend on how the prompt is designed, which eventually affects the meaning of the generated text \cite{lu2021fantastically}. In terms of models, especially in the case of GPT-3 models, we were somewhat limited to the number of versions that we could experiment with due to limited computational resources and accessing it as a paid service. Recent versions of the ChatGPT model as well as more open source models could also be studied. Additionally, given the non-deterministic nature of text generations, multiple $S3'$ continuations could be generated and studied. Although this paper focused primarily on higher-resource languages within the same language family, we plan to extend the inquiry to include lower-resource languages from different language families.


\section*{Ethics Consideration}

The use of idiomatic expressions in natural language can potentially alter the intended meaning of a message. If a language model is unable to accurately interpret these idiomatic expressions, it can easily lead to a misinterpretation of the message and negatively impact the overall effectiveness of the model. Language models have also been shown to contain gender biases \cite{lucy2021gender}. As we used existing datasets from credible sources (SemEval 2022, Task 2) in our experiments, we did not verify every instance manually but considering that the data originated from `naturally occurring sentences', it is possible that the data may contain unintended biases or offensive content.


\section*{Limitations}
We explored only a handful of prompts in this work. There are several ways in any language to write the same prompt. As such, the generated text might depend on how the prompt is designed eventually affecting the meaning of the generated text \cite{lu2021fantastically}. Another limitation of our work is that human assessment was only conducted on English samples.  In terms of models, especially in the case of GPT-3 models, we were limited to the number of variants we could experiment with due to limited computational resources and accessing it as a paid service.

\section*{Acknowledgments}


We would like to thank the anonymous reviewers and the PortNLP research group for their insightful feedback. This research was supported by the National Science Foundation under Grant No. CRII:RI-2246174.

\bibliography{anthology,custom}

\begin{thebibliography}{25}
\expandafter\ifx\csname natexlab\endcsname\relax\def\natexlab#1{#1}\fi

\bibitem[{Banerjee and Lavie(2005)}]{banerjee2005meteor}
Satanjeev Banerjee and Alon Lavie. 2005.
\newblock Meteor: An automatic metric for mt evaluation with improved correlation with human judgments.
\newblock In \emph{Proceedings of the acl workshop on intrinsic and extrinsic evaluation measures for machine translation and/or summarization}, pages 65--72.

\bibitem[{Bird(2006)}]{bird2006nltk}
Steven Bird. 2006.
\newblock Nltk: the natural language toolkit.
\newblock In \emph{Proceedings of the COLING/ACL 2006 Interactive Presentation Sessions}, pages 69--72.

\bibitem[{Chakrabarty et~al.(2022)Chakrabarty, Choi, and Shwartz}]{chakrabarty2022s}
Tuhin Chakrabarty, Yejin Choi, and Vered Shwartz. 2022.
\newblock It’s not rocket science: Interpreting figurative language in narratives.
\newblock \emph{Transactions of the Association for Computational Linguistics}, 10:589--606.

\bibitem[{Chakrabarty et~al.(2021)Chakrabarty, Ghosh, Poliak, and Muresan}]{chakrabarty-etal-2021-figurative}
Tuhin Chakrabarty, Debanjan Ghosh, Adam Poliak, and Smaranda Muresan. 2021.
\newblock \href {https://doi.org/10.18653/v1/2021.findings-acl.297} {Figurative language in recognizing textual entailment}.
\newblock In \emph{Findings of the Association for Computational Linguistics: ACL-IJCNLP 2021}, pages 3354--3361, Online. Association for Computational Linguistics.

\bibitem[{Dashtipour et~al.(2022)Dashtipour, Gogate, Gelbukh, and Hussain}]{dashtipour2022extending}
Kia Dashtipour, Mandar Gogate, Alexander Gelbukh, and Amir Hussain. 2022.
\newblock Extending persian sentiment lexicon with idiomatic expressions for sentiment analysis.
\newblock \emph{Social Network Analysis and Mining}, 12(1):1--13.

\bibitem[{Fadaee et~al.(2018)Fadaee, Bisazza, and Monz}]{fadaee2018examining}
Marzieh Fadaee, Arianna Bisazza, and Christof Monz. 2018.
\newblock Examining the tip of the iceberg: A data set for idiom translation.
\newblock \emph{arXiv preprint arXiv:1802.04681}.

\bibitem[{He et~al.(2021)He, Liu, Gao, and Chen}]{he2021deberta}
Pengcheng He, Xiaodong Liu, Jianfeng Gao, and Weizhu Chen. 2021.
\newblock \href {https://openreview.net/forum?id=XPZIaotutsD} {Deberta: Decoding-enhanced bert with disentangled attention}.
\newblock In \emph{International Conference on Learning Representations}.

\bibitem[{Jhamtani et~al.(2021)Jhamtani, Gangal, Hovy, and Berg-Kirkpatrick}]{jhamtani2021investigating}
Harsh Jhamtani, Varun Gangal, Eduard Hovy, and Taylor Berg-Kirkpatrick. 2021.
\newblock Investigating robustness of dialog models to popular figurative language constructs.
\newblock \emph{arXiv preprint arXiv:2110.00687}.

\bibitem[{Joshi et~al.(2020{\natexlab{a}})Joshi, Santy, Budhiraja, Bali, and Choudhury}]{joshietal2020state}
Pratik Joshi, Sebastin Santy, Amar Budhiraja, Kalika Bali, and Monojit Choudhury. 2020{\natexlab{a}}.
\newblock \href {https://doi.org/10.18653/v1/2020.acl-main.560} {The state and fate of linguistic diversity and inclusion in the {NLP} world}.
\newblock In \emph{Proceedings of the 58th Annual Meeting of the Association for Computational Linguistics}.

\bibitem[{Joshi et~al.(2020{\natexlab{b}})Joshi, Santy, Budhiraja, Bali, and Choudhury}]{joshi-etal-2020-state}
Pratik Joshi, Sebastin Santy, Amar Budhiraja, Kalika Bali, and Monojit Choudhury. 2020{\natexlab{b}}.
\newblock \href {https://doi.org/10.18653/v1/2020.acl-main.560} {The state and fate of linguistic diversity and inclusion in the {NLP} world}.
\newblock In \emph{Proceedings of the 58th Annual Meeting of the Association for Computational Linguistics}, pages 6282--6293, Online. Association for Computational Linguistics.

\bibitem[{Korkontzelos et~al.(2013)Korkontzelos, Zesch, Zanzotto, and Biemann}]{korkontzelos2013semeval}
Ioannis Korkontzelos, Torsten Zesch, Fabio~Massimo Zanzotto, and Chris Biemann. 2013.
\newblock Semeval-2013 task 5: Evaluating phrasal semantics.
\newblock In \emph{Second Joint Conference on Lexical and Computational Semantics (* SEM), Volume 2: Proceedings of the Seventh International Workshop on Semantic Evaluation (SemEval 2013)}, pages 39--47.

\bibitem[{Lin(2004)}]{lin2004rouge}
Chin-Yew Lin. 2004.
\newblock Rouge: A package for automatic evaluation of summaries.
\newblock In \emph{Text summarization branches out}, pages 74--81.

\bibitem[{Liu et~al.(2022)Liu, Cui, Zheng, and Neubig}]{liu2022testing}
Emmy Liu, Chen Cui, Kenneth Zheng, and Graham Neubig. 2022.
\newblock Testing the ability of language models to interpret figurative language.
\newblock \emph{arXiv preprint arXiv:2204.12632}.

\bibitem[{Lu et~al.(2021)Lu, Bartolo, Moore, Riedel, and Stenetorp}]{lu2021fantastically}
Yao Lu, Max Bartolo, Alastair Moore, Sebastian Riedel, and Pontus Stenetorp. 2021.
\newblock Fantastically ordered prompts and where to find them: Overcoming few-shot prompt order sensitivity.
\newblock \emph{arXiv preprint arXiv:2104.08786}.

\bibitem[{Lucy and Bamman(2021)}]{lucy2021gender}
Li~Lucy and David Bamman. 2021.
\newblock Gender and representation bias in gpt-3 generated stories.
\newblock In \emph{Proceedings of the Third Workshop on Narrative Understanding}, pages 48--55.

\bibitem[{Moussallem et~al.(2018)Moussallem, Sherif, Esteves, Zampieri, and Ngomo}]{moussallem2018lidioms}
Diego Moussallem, Mohamed~Ahmed Sherif, Diego Esteves, Marcos Zampieri, and Axel-Cyrille~Ngonga Ngomo. 2018.
\newblock Lidioms: A multilingual linked idioms data set.
\newblock \emph{arXiv preprint arXiv:1802.08148}.

\bibitem[{Peng et~al.(2015)Peng, Feldman, and Jazmati}]{peng2015classifying}
Jing Peng, Anna Feldman, and Hamza Jazmati. 2015.
\newblock Classifying idiomatic and literal expressions using vector space representations.
\newblock In \emph{Proceedings of the International Conference Recent Advances in Natural Language Processing}, pages 507--511.

\bibitem[{Tan and Jiang(2021)}]{tan2021does}
Minghuan Tan and Jing Jiang. 2021.
\newblock Does bert understand idioms? a probing-based empirical study of bert encodings of idioms.
\newblock In \emph{Proceedings of the International Conference on Recent Advances in Natural Language Processing (RANLP 2021)}, pages 1397--1407.

\bibitem[{Tang(2022)}]{tang2022petci}
Kenan Tang. 2022.
\newblock Petci: A parallel english translation dataset of chinese idioms.
\newblock \emph{arXiv preprint arXiv:2202.09509}.

\bibitem[{Tayyar~Madabushi et~al.(2021)Tayyar~Madabushi, Gow-Smith, Scarton, and Villavicencio}]{tayyar2021astitchinlanguagemodels}
Harish Tayyar~Madabushi, Edward Gow-Smith, Carolina Scarton, and Aline Villavicencio. 2021.
\newblock \href {https://doi.org/10.18653/v1/2021.findings-emnlp.294} {{AS}titch{I}n{L}anguage{M}odels: Dataset and methods for the exploration of idiomaticity in pre-trained language models}.
\newblock In \emph{Findings of the Association for Computational Linguistics: EMNLP 2021}, pages 3464--3477, Punta Cana, Dominican Republic. Association for Computational Linguistics.

\bibitem[{Tedeschi et~al.(2022)Tedeschi, Martelli, and Navigli}]{tedeschi2022id10m}
Simone Tedeschi, Federico Martelli, and Roberto Navigli. 2022.
\newblock Id10m: Idiom identification in 10 languages.
\newblock In \emph{Findings of the Association for Computational Linguistics: NAACL 2022}, pages 2715--2726.

\bibitem[{Tedeschi and Navigli(2022)}]{tedeschi2022ner4id}
Simone Tedeschi and Roberto Navigli. 2022.
\newblock Ner4id at semeval-2022 task 2: Named entity recognition for idiomaticity detection.
\newblock In \emph{Proceedings of the 16th International Workshop on Semantic Evaluation (SemEval-2022). Association for Computational Linguistics}.

\bibitem[{Zhang et~al.(2019)Zhang, Kishore, Wu, Weinberger, and Artzi}]{zhang2019bertscore}
Tianyi Zhang, Varsha Kishore, Felix Wu, Kilian~Q Weinberger, and Yoav Artzi. 2019.
\newblock Bertscore: Evaluating text generation with bert.
\newblock \emph{arXiv preprint arXiv:1904.09675}.

\bibitem[{Zheng et~al.(2019)Zheng, Huang, and Sun}]{zheng2019chid}
Chujie Zheng, Minlie Huang, and Aixin Sun. 2019.
\newblock Chid: A large-scale chinese idiom dataset for cloze test.
\newblock \emph{arXiv preprint arXiv:1906.01265}.

\bibitem[{Zhou et~al.(2021)Zhou, Zeng, Gong, and Bhat}]{zhou2022idiomatic}
Jianing Zhou, Ziheng Zeng, Hongyu Gong, and Suma Bhat. 2021.
\newblock \href {http://arxiv.org/abs/2112.08592} {Idiomatic expression paraphrasing without strong supervision}.
\newblock \emph{CoRR}, abs/2112.08592.

\end{thebibliography}
\bibliographystyle{acl_natbib}

\appendix



\end{document}